\documentclass[10pt]{article}

\usepackage[margin=1in]{geometry}
\usepackage{amsmath,amssymb,amsfonts}
\usepackage{graphicx}
\usepackage{algorithm,algorithmic}
\usepackage{hyperref}
\usepackage{xcolor}
\usepackage{setspace}
\title{\textbf{Predicting Treatment Response in Body Dysmorphic Disorder with Interpretable Machine Learning}}
\author{Omar Costilla Reyes, MIT \\ Morgan Talbot, MIT and Harvard}
\date{}

\begin{document}

\maketitle

\begin{abstract}
\noindent
Body Dysmorphic Disorder (BDD) is a highly prevalent and frequently underdiagnosed condition characterized by persistent, intrusive preoccupations with perceived defects in physical appearance. In this extended analysis, we employ multiple machine learning approaches to predict treatment outcomes -- specifically treatment response and remission -- with an emphasis on interpretability to ensure clinical relevance and utility. Across the various models investigated, treatment credibility emerged as the most potent predictor, surpassing traditional markers such as baseline symptom severity or comorbid conditions. Notably, while simpler models (e.g., logistic regression and support vector machines) achieved competitive predictive performance, decision tree analyses provided unique insights by revealing clinically interpretable threshold values in credibility scores. These thresholds can serve as practical guideposts for clinicians when tailoring interventions or allocating treatment resources. We further contextualize our findings within the broader literature on BDD, addressing technology-based therapeutics, digital interventions, and the psychosocial determinants of treatment engagement. An extensive array of references situates our results within current research on BDD prevalence, suicidality risks, and digital innovation. Our work underscores the potential of integrating rigorous statistical methodologies with transparent machine learning models. By systematically identifying modifiable predictors -- such as treatment credibility -- we propose a pathway toward more targeted, personalized, and ultimately efficacious interventions for individuals with BDD.
\end{abstract}

\section{Introduction}

Body Dysmorphic Disorder (BDD) affects an estimated 1.7\% to 2.9\% of the general population \cite{Buhlmann2010, Koran2008}, with even higher prevalence in specialized settings such as dermatology and cosmetic surgery clinics. BDD is characterized by persistent, distressing concerns about perceived defects in physical appearance that are often imperceptible or negligible to others \cite{Phillips1997, Phillips2005, Schieber2015}. Numerous studies have demonstrated that individuals with BDD frequently suffer from a diminished quality of life \cite{Phillips2000, Marques2011} and an elevated risk of suicidality \cite{Angelakis2016}. Despite the profound impact on functional impairment and overall well-being, BDD remains underdiagnosed in many clinical environments.

Cognitive behavioral therapy (CBT) and selective serotonin reuptake inhibitors (SSRIs) are recognized as first-line treatments for BDD \cite{Phillips2021, Wilhelm2019, Harrison2016}. Although these interventions are effective for a substantial subset of patients, a considerable number of individuals do not achieve full remission or experience relapse following initial improvements \cite{Weingarden2021, Hogg2023}. In addition to the inherent variability in treatment response, numerous barriers—including stigma, scarcity of specialized treatment providers, and an incomplete understanding of BDD’s etiology—complicate the delivery of effective care \cite{Marques2011, Yousaf2015}.

The rapid expansion of digital health solutions offers new avenues for delivering mental health interventions, particularly for populations with limited access to traditional care. App-based and internet-delivered CBT interventions for BDD have shown promising results \cite{Weingarden2020, Enander2016, Wilhelm2022}, suggesting that technology can mitigate barriers related to accessibility and stigma. Concurrently, machine learning (ML) techniques have the potential to augment these digital approaches by predicting which individuals may require more intensive or specialized interventions. Such predictive capabilities could reduce dropout rates and optimize treatment outcomes by guiding clinicians in personalizing therapeutic approaches \cite{Greenberg2022, Flygare2020}. However, for ML-based predictions to be adopted in clinical practice, the underlying models must be both transparent and interpretable \cite{molnar2019interpretable}.

While non-linear models such as random forests and gradient boosting machines frequently yield high accuracy, they often function as ``black boxes,'' thus limiting clinicians’ confidence in their practical application. In contrast, linear models (e.g., logistic regression and linear support vector machines) and simpler tree-based methods (e.g., decision trees) provide rule-based structures that are more accessible and understandable. This interpretability is especially critical in the mental health domain, where patient beliefs and treatment credibility play a decisive role in shaping therapy engagement and outcomes \cite{Devilly2000, Arch2015, Constantino2018, Bernstein2023}. \\

\textbf{Study Aims and Organization}
The primary aims of this study are to: 
\begin{enumerate}
    \item Compare the predictive performance of several ML models in forecasting treatment response and remission among patients with BDD.
    \item Emphasize the role of model interpretability in directly informing clinical decision-making.
    \item Investigate the pivotal role of treatment credibility as a modifiable predictor of treatment outcomes.
\end{enumerate}
Our methodological framework incorporates multiple imputation \cite{Rubin1976, Rubin2004}, rigorous cross-validation procedures \cite{James2013}, and forward feature selection to mitigate overfitting and enhance model clarity. Throughout the manuscript, we integrate a comprehensive set of references spanning BDD epidemiology, digital health innovations, and advanced ML applications to contextualize our findings within the broader literature.

\section{Methods}
\textbf{Participant Recruitment and Data Collection}
Participants were recruited from outpatient and community settings as part of a clinical trial investigating CBT for BDD \cite{Wilhelm2019, Phillips2021}. Eligible participants met DSM-5 criteria for BDD; comorbid conditions were permitted provided they did not pose an acute risk or induce severe clinical instability (e.g., active psychosis, imminent suicidality). Demographic and clinical data were obtained through structured clinical interviews (including administration of the MINI \cite{Sheehan1998}) and validated self-report questionnaires. All data were managed using a secure database system to ensure confidentiality and data integrity \cite{Harris2009}.

\paragraph{Outcome Measures.}
The study focused on two binary outcomes:
\begin{enumerate}
    \item \textbf{Treatment Response}: Defined as at least a 30\% reduction in BDD-YBOCS scores from baseline, consistent with previous studies \cite{Greenberg2019, Greenberg2022}.
    \item \textbf{Remission}: Defined as a final BDD-YBOCS score $\leq 16$, in line with empirically established thresholds \cite{Fernandez2021}.
\end{enumerate}

\textbf{Candidate Predictors (Features)}
Based on extant literature \cite{Harrison2016, Phillips2021, Hogg2023}, we included 17 candidate predictors encompassing demographic, clinical, and treatment-related factors:
\begin{itemize}
    \item \textbf{Demographics}: 
    \begin{itemize}
        \item Age 
        \item Gender identity (dichotomized as female vs. male/other)
        \item Sexual minority status (dichotomized as heterosexual vs. non-heterosexual), in recognition of unique barriers faced by LGBTIQ+ individuals \cite{Bowen2016, Gilbey2020, Lucassen2013}
        \item Race/ethnicity (non-Hispanic White vs. all others), acknowledging documented disparities in mental healthcare access and outcomes \cite{McGuire2008}
        \item Education status (e.g., high school completion versus post-graduate education)
    \end{itemize}
    \item \textbf{Baseline Clinical Questionnaires}:
    \begin{itemize}
        \item BDD-YBOCS \cite{Phillips1997, Phillips2014} 
        \item URICA (assessing readiness to change) \cite{McConnaughy1983}
        \item BABS (evaluating insight and delusional beliefs) \cite{Eisen1998, Phillips2013}
        \item QIDS for depression severity \cite{Rush2003}
        \item Treatment credibility and expectancy scales \cite{Devilly2000, Arch2015, Constantino2018}
    \end{itemize}
    \item \textbf{Other Variables}: 
    \begin{itemize}
        \item Duration of BDD illness
        \item Treatment group assignment (e.g., standard CBT vs. enhanced or technology-assisted CBT)
        \item SSRI usage at baseline (binary)
        \item Presence of any comorbid psychiatric disorder (binary)
        \item Self-reported impact of the COVID-19 pandemic on BDD symptoms
    \end{itemize}
\end{itemize}

\textbf{Machine Learning Models}
We evaluated five machine learning modeling approaches using the SciKit-Learn library \cite{Pedregosa2011}:
\begin{enumerate}
    \item \textbf{Logistic Regression (LR)}
    \item \textbf{Support Vector Machine (SVM)} with a linear kernel
    \item \textbf{K-Nearest Neighbors (KNN)}
    \item \textbf{Decision Trees}
    \item \textbf{Random Forests}
\end{enumerate}

\textbf{Multiple Imputation and Cross-Validation}
Missing data in both outcome and predictor variables were addressed through multiple imputation \cite{Rubin1976, Rubin2004}, resulting in 100 imputed datasets. Each dataset was further analyzed using 5-fold cross-validation, yielding a total of 500 training-testing iterations. Model performance was assessed primarily using the area under the receiver operating characteristic curve (AUC).

\textbf{Forward Feature Selection}
To minimize overfitting while preserving interpretability, forward feature selection was implemented with a strict requirement: each newly added predictor had to improve the AUC by at least 0.05. This stringent criterion yielded parsimonious models that are not only robust but also readily interpretable—a key advantage in clinical decision-making contexts \cite{Hosmer2013, James2013}.

\section{Results}

Table \ref{tab:ml_performance} presents the mean AUC values and the final set of features selected (in order of selection) across the five models for both predicting treatment response and remission. Although random forests achieved competitive performance, linear models such as logistic regression and SVM generally matched or exceeded their accuracy. While decision trees demonstrated slightly lower AUC values—particularly for remission—they offered critical interpretative insights that can guide clinical application.

\begin{figure}[!ht]
\centering
\includegraphics[width=0.75\columnwidth]{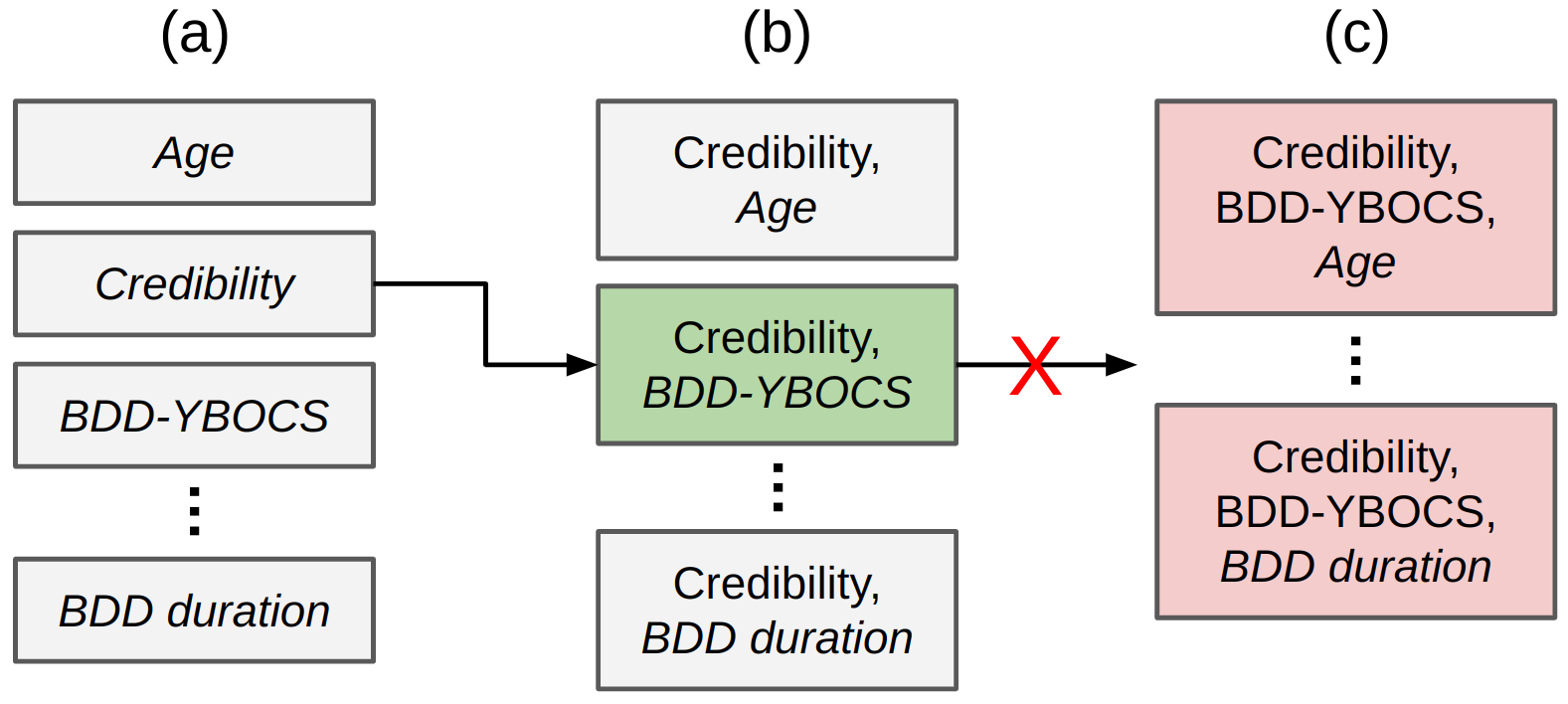}
\caption{\small Illustration of the forward feature selection procedure used in our study. In (a), candidate features (e.g., Age, Credibility, BDD-YBOCS) are initially considered. In (b), the addition of BDD duration fails to meet the 0.05 AUC-improvement threshold (indicated by the red cross). In (c), the final selected combination of features (shown in pink) includes Credibility and BDD-YBOCS.}
\label{fig:forward_selection}
\end{figure}

\begin{table}[!ht]
\centering
\begin{tabular}{lll}
\hline
\multicolumn{3}{c}{\textbf{(a) Predicting Treatment Response}}\\
\hline
\textbf{Model} & \textbf{AUC} & \textbf{Selected Features (Order)} \\
\hline
Logistic Regression     & 0.726  & credibility, sexual\_minority  \\
SVM                     & \textbf{0.733}  & credibility, sexual\_minority, gender\_identity, babs\_tot\_recalc \\
Decision Tree           & 0.696  & credibility, gender\_identity  \\
Random Forest           & 0.723  & credibility, gender\_identity  \\
KNN                     & 0.682  & credibility, gender\_identity  \\
\hline
\\
\hline
\multicolumn{3}{c}{\textbf{(b) Predicting Illness Remission}} \\
\hline
\textbf{Model} & \textbf{AUC} & \textbf{Selected Features (Order)}        \\
\hline
Logistic Regression     & 0.697  & credibility \\
SVM                     & \textbf{0.718}  & credibility, BDD-YBOCS   \\
Decision Tree           & 0.616  & credibility \\
Random Forest           & 0.709  & credibility \\
KNN                     & 0.688  & credibility, postgrad\_education \\
\hline
\end{tabular}
\caption{\small{Average AUC values from 500 cross-validation runs (5-fold $\times$ 100 imputations) for (a) treatment response and (b) remission. Models employed forward feature selection; features are listed in the order of their selection.}}
\label{tab:ml_performance}
\end{table}

Across all models, \textbf{treatment credibility} was the first feature selected, underscoring its substantial impact on predictive performance for both treatment response and remission. This observation is consistent with prior research demonstrating that patients’ beliefs and perceptions about therapy significantly influence engagement and clinical outcomes \cite{Bernstein2023, Constantino2018}. Interestingly, although treatment expectancy was also measured, its inclusion in the final models was diminished once credibility was accounted for, likely due to the high correlation between these constructs ($r \approx 0.67$) \cite{Devilly2000, Arch2015}.

Although decision trees yielded marginally lower AUC values for remission, they provided clear, interpretable decision thresholds that can be directly applied in clinical settings. In particular, decision trees consistently identified scores of \textbf{16} and \textbf{22} on the 3-item Likert-based credibility scale (ranging from 3 to 27) as critical cutoffs. Patients scoring $\leq 16$ were generally less likely to respond or achieve remission, while those scoring $> 22$ were predominantly successful in their treatment outcomes (see Table \ref{tab:credibility_odds_ratio}). Such threshold values offer a simple heuristic for clinicians to stratify patients based on their likelihood of treatment success.

\begin{figure}[!ht]
\centering
\includegraphics[width=0.75\columnwidth]{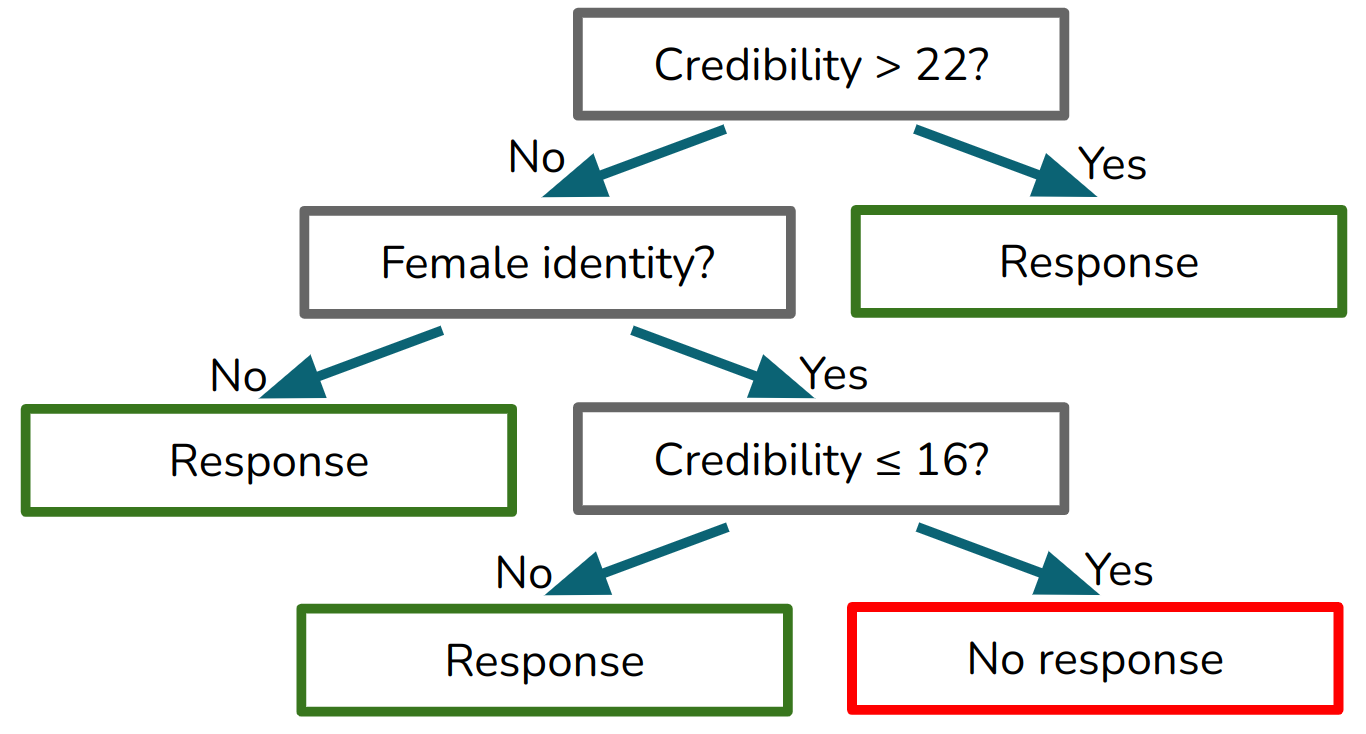}
\caption{\small A representative decision tree for predicting treatment response, highlighting how treatment credibility interacts with gender identity. Patients scoring above 22 on credibility are likely to respond; in contrast, female-identifying patients with credibility scores below 16 are at higher risk for non-response.}
\label{fig:gender_decision_tree}
\end{figure}

\begin{figure}[!ht]
\centering
\includegraphics[width=0.75\columnwidth]{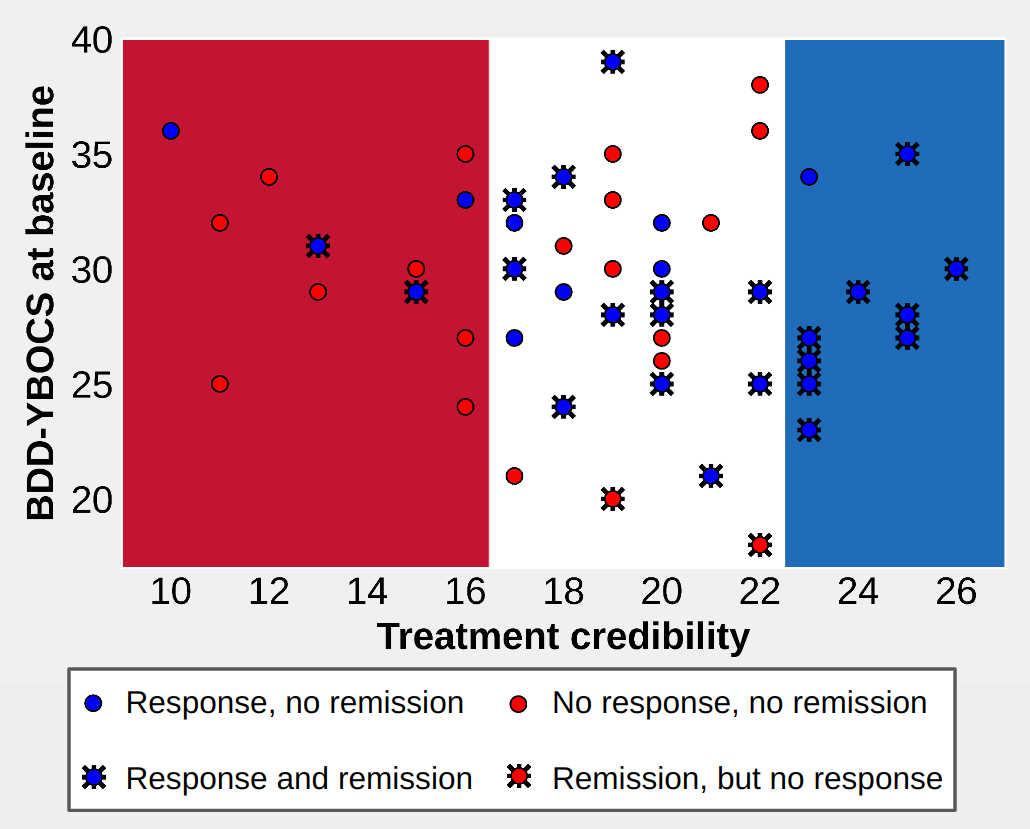}
\caption{\small Scatter plot of BDD patients by baseline treatment credibility (x-axis) and BDD-YBOCS (y-axis). Blue markers denote treatment responders; red markers denote non-responders; star-like markers indicate remission. Patients with credibility scores $\leq 16$ rarely responded or remitted, whereas scores $> 22$ were strongly predictive of positive outcomes.}
\label{fig:credibility_scatter}
\end{figure}

\begin{table}[!ht]
\centering
\begin{tabular}{lcc}
\hline
\multicolumn{3}{c}{\textbf{Credibility Score Thresholds and Odds Ratios}} \\
\hline
\textbf{Threshold} & \textbf{Outcome} & \textbf{OR (p-value)}       \\
\hline
$\leq 16$  & Response & 0.315 (p = 0.04) \\
           & Remission & 0.324 (p = 0.06)\\
$> 22$     & Response & $> 7.61$ (p $< 0.044$)   \\
           & Remission & 11.0 (p = 0.007) \\
\hline
\end{tabular}
\caption{\small Odds ratios for treatment response and remission based on credibility score thresholds. Decision tree analyses consistently identified these cut points as clinically relevant.}
\label{tab:credibility_odds_ratio}
\end{table}

Secondary predictors emerged in several models, with gender identity being a recurrent feature in predicting treatment response. Notably, female-identifying patients exhibiting low credibility scores were often associated with minimal symptom reduction. Although prior studies have noted that men may delay help-seeking due to social stigma \cite{Yousaf2015}, our findings indicate that, in the context of BDD, the interaction between female gender identity and skepticism regarding therapy correlates strongly with poorer outcomes. Additional variables, such as baseline BDD-YBOCS severity and higher education levels, were intermittently selected, suggesting their potential to refine predictive accuracy in certain patient subgroups \cite{Greenberg2019, Flygare2020}.

\section{Discussion}

Our analysis confirms that in small-sample ML applications within psychiatry, the performance gap between sophisticated non-linear models (e.g., random forests) and simpler, interpretable models (e.g., logistic regression and SVM) is relatively modest \cite{Flygare2020, Greenberg2022}. This finding emphasizes that model interpretability need not be sacrificed for predictive accuracy. For disorders like BDD—where the clinician-patient relationship and treatment transparency are paramount—transparent models may facilitate better integration into routine practice \cite{molnar2019interpretable, Hosmer2013}.

The central role of treatment credibility observed in our study reinforces the notion that patient perceptions are pivotal in determining clinical outcomes. Previous research has consistently linked higher credibility and expectancy scores with improved engagement and adherence in CBT \cite{Constantino2018, Arch2015}. In the context of BDD, preliminary investigations into smartphone-delivered CBT have similarly underscored the importance of initial treatment perceptions \cite{Bernstein2023, Weingarden2020}. A key implication of these findings is the potential to develop pre-treatment interventions—such as brief educational modules or motivational interviewing—that specifically aim to elevate treatment credibility. Achieving scores above critical thresholds (e.g., surpassing 22) may enhance adherence to treatment protocols, including between-session assignments and exposure tasks.

The rapid proliferation of digital mental health interventions underscores the need to consider how treatment credibility interacts with technology acceptance. For instance, sexual and gender minority groups may experience unique challenges related to the acceptance and perceived legitimacy of novel therapeutic formats \cite{Bowen2016, Lucassen2013}. Tailoring psychoeducational content to accommodate cultural and personal contexts is essential in ensuring equitable access and engagement \cite{McGuire2008}. Digital platforms, by mitigating geographical barriers and reducing stigma, offer a promising avenue for delivering personalized interventions. Integrating ML-driven diagnostics within these platforms could further enable the identification of patients at risk of dropout or non-response, thereby allowing for the timely implementation of targeted interventions.

\paragraph{A Simple Heuristic Based on Credibility.}
Clinicians can utilize the credibility thresholds identified through decision tree analysis (approximately 16 and 22 on a 27-point scale) as a rapid screening tool:
\begin{itemize}
    \item \textbf{Scores $\leq 16$:} These patients may benefit from intensified psychoeducation, strategies to address misconceptions, and motivational interviewing techniques either before or concurrently with standard BDD interventions.
    \item \textbf{Scores $> 22$:} These individuals exhibit a higher likelihood of positive outcomes; reinforcing their confidence and sustaining engagement could further enhance treatment efficacy.
\end{itemize}

\paragraph{Integration with Standard Measures.}
While established screening instruments such as the BDD-YBOCS remain indispensable, incorporating treatment credibility thresholds as adjunctive markers can inform decisions regarding treatment intensity and modality. For example, patients exhibiting low credibility scores might initially undergo a more supportive, trust-building phase prior to the commencement of full-scale CBT \cite{Arch2015, Devilly2000}.

\paragraph{Future Directions for Practice.}
Clinicians exploring technology-assisted treatments might embed credibility enhancing components into early digital modules of app-based or telehealth CBT. Personalized content that resonates with patients’ cultural backgrounds or success narratives from demographically similar individuals could effectively mitigate skepticism and improve adherence. Further empirical research is needed to confirm whether augmenting initial treatment credibility translates into sustained improvements in therapy outcomes.

Earlier studies by Flygare et al. \cite{Flygare2020} and Hogg et al. \cite{Hogg2023} have employed non-linear models and systematic reviews to identify predictors of remission in BDD. Our study extends these findings in two critical ways:
\begin{enumerate}
    \item We applied a forward feature selection strategy with a rigorous 0.05 AUC-improvement criterion, thereby ensuring a minimal yet highly interpretable set of predictors.
    \item We elucidated specific credibility thresholds, as identified by decision trees, which enhance the clinical applicability of our findings.
\end{enumerate}
These observations align with previous reports emphasizing the central role of treatment expectancy and credibility in optimizing exposure therapy for body image concerns \cite{Arch2015}.

\paragraph{Sample Size and External Validity.}
As with many BDD studies \cite{Harrison2016, Greenberg2019}, our sample size was relatively modest, which may limit the generalizability of our findings. Future multi-site studies with larger, more diverse populations are needed to confirm whether the prognostic value of treatment credibility extends to broader clinical settings.

\paragraph{Reliance on Self-Reported Data.}
The reliance on self-reported measures for treatment credibility, expectancy, and clinical outcomes may introduce bias or variability related to transient mood states \cite{Devilly2000}. Complementary qualitative methods or repeated assessments could provide deeper insight into the stability of these constructs over time.

\paragraph{Causal Ambiguities.}
While our findings indicate a strong association between treatment credibility and positive outcomes, they do not establish a causal relationship. Randomized, controlled trials specifically targeting low-credibility patients with pre-treatment psychoeducational interventions would be valuable in determining whether elevating credibility directly improves treatment adherence and clinical response.

\paragraph{Role of Digital Interventions and Technology Acceptance.}
Although our study highlights the potential synergy between ML-driven insights and digital health technologies, further research is warranted to explore user acceptance, engagement longevity, and the impact of culturally tailored digital interventions—particularly among vulnerable populations such as LGBTIQ+ youth \cite{Lucassen2013, Gilbey2020}.

\section{Conclusions}
This expanded study reinforces the utility of interpretable machine learning models in predicting treatment response and remission in individuals with Body Dysmorphic Disorder. The key findings include:
\begin{enumerate}
    \item \textbf{Treatment credibility} is a predominant predictor, outperforming many classical clinical and demographic variables.
    \item Simpler, interpretable models (e.g., logistic regression and linear SVM) often rival more complex approaches in small-sample settings, demonstrating that transparency can be maintained without sacrificing predictive accuracy.
    \item Decision tree analyses have identified specific \textbf{credibility thresholds} (approximately 16 and 22 on a 27-point scale) that are directly applicable in tailoring therapeutic strategies.
\end{enumerate}

These insights strongly resonate with the broader mental health literature, emphasizing that enhancing patients’ perceptions of treatment can meaningfully improve therapy engagement and outcomes \cite{Constantino2018, Bernstein2023}. Interventions aimed at boosting initial treatment credibility—whether through targeted psychoeducation, motivational interviewing, or personalized digital content—could potentially amplify the benefits of CBT, particularly among patients predisposed to skepticism. By merging robust statistical techniques (e.g., multiple imputation, cross-validation) with user-friendly, interpretable models, our approach offers a promising template for advancing personalized mental healthcare. Future research should expand on these findings by increasing sample sizes, exploring causative mechanisms, and integrating ML-driven insights into both in-person and digital therapeutic platforms.

\bibliography{references}  

\begin{thebibliography}{10}

\bibitem{Buhlmann2010}
U.~Buhlmann, H.~Glaesmer, R.~Mewes, J.~M. Fama, S.~Wilhelm, E.~Br{\"a}hler, and W.~Rief.
\newblock Updates on the prevalence of body dysmorphic disorder: A population-based survey.
\newblock {\em Psychiatry Research}, 178:171--175, 2010.

\bibitem{Koran2008}
L.~M. Koran, E.~Abujaoude, M.~D. Large, and R.~T. Serpe.
\newblock The prevalence of body dysmorphic disorder in the united states adult population.
\newblock {\em CNS Spectrums}, 13:316--322, 2008.

\bibitem{Phillips1997}
K.~A. Phillips, E.~Hollander, S.~A. Rasmussen, B.~R. Aronowitz, C.~DeCaria, and W.~K. Goodman.
\newblock A severity rating scale for body dysmorphic disorder: Development, reliability, and validity of a modified version of the yale-brown obsessive compulsive scale.
\newblock {\em Psychopharmacology Bulletin}, 33:17--22, 1997.

\bibitem{Phillips2005}
K.~A. Phillips, W.~Menard, C.~Fay, and R.~Weisberg.
\newblock Demographic characteristics, phenomenology, comorbidity, and family history in 200 individuals with body dysmorphic disorder.
\newblock {\em Psychosomatics}, 46:317--325, 2005.

\bibitem{Schieber2015}
K.~Schieber, I.~Kollei, M.~de~Zwaan, and A.~Martin.
\newblock Classification of body dysmorphic disorder - what is the advantage of the new dsm-5 criteria?
\newblock {\em Journal of Psychosomatic Research}, 78:223--227, 2015.

\bibitem{Phillips2000}
K.~A. Phillips.
\newblock Quality of life for patients with body dysmorphic disorder.
\newblock {\em Journal of Nervous and Mental Disease}, 188:1703--1707, 2000.

\bibitem{Marques2011}
L.~Marques, H.~M. Weingarden, N.~J. Leblanc, and S.~Wilhelm.
\newblock Treatment utilization and barriers to treatment engagement among people with body dysmorphic symptoms.
\newblock {\em Journal of Psychosomatic Research}, 70:286--293, 2011.

\bibitem{Angelakis2016}
I.~Angelakis, P.~A. Gooding, and M.~Panagioti.
\newblock Suicidality in body dysmorphic disorder (bdd): A systematic review with meta-analysis.
\newblock {\em Clinical Psychology Review}, 49:55--66, 2016.

\bibitem{Phillips2021}
K.~A. Phillips, J.~L. Greenberg, S.~S. Hoeppner, H.~Weingarden, S.~O'Keefe, A.~Keshaviah, ..., and S.~Wilhelm.
\newblock Predictors and moderators of symptom change during cognitive-behavioral therapy or supportive psychotherapy for body dysmorphic disorder.
\newblock {\em Journal of Affective Disorders}, 287:34--40, 2021.
\newblock PMCID: PMC8276884.

\bibitem{Wilhelm2019}
S.~Wilhelm, K.~A. Phillips, J.~L. Greenberg, S.~M. O’Keefe, S.~S. Hoeppner, A.~Keshaviah, and D.~A. Schoenfeld.
\newblock Efficacy and posttreatment effects of therapist-delivered cognitive behavioral therapy vs supportive psychotherapy for adults with body dysmorphic disorder: A randomized clinical trial.
\newblock {\em JAMA Psychiatry}, 76:363--373, 2019.
\newblock PMCID: PMC6450292.

\bibitem{Harrison2016}
A.~Harrison, L.~Fern\'andez de~la Cruz, J.~Enander, J.~Radua, and D.~Mataix-Cols.
\newblock Cognitive-behavioral therapy for body dysmorphic disorder: A systematic review and meta-analysis of randomized controlled trials.
\newblock {\em Clinical Psychology Review}, 48:433--451, 2016.

\bibitem{Weingarden2021}
H.~Weingarden, S.~S. Hoeppner, I.~Snorrason, J.~L. Greenberg, K.~A. Phillips, and S.~Wilhelm.
\newblock Rates of remission, sustained remission, and recurrence in a randomized controlled trial of cognitive behavioral therapy versus supportive psychotherapy for body dysmorphic disorder.
\newblock {\em Depression and Anxiety}, 2021.
\newblock PMCID: PMC8443701.

\bibitem{Hogg2023}
E.~Hogg, P.~Adamopoulos, and G.~Krebs.
\newblock Predictors and moderators of treatment response in cognitive behavioural therapy for body dysmorphic disorder: A systematic review.
\newblock {\em Journal of Obsessive-Compulsive and Related Disorders}, 38, 2023.

\bibitem{Yousaf2015}
O.~Yousaf, E.~A. Grunfeld, and M.~S. Hunter.
\newblock A systematic review of the factors associated with delays in medical and psychological help-seeking among men.
\newblock {\em Health Psychology Review}, 9:264--276, 2015.

\bibitem{Weingarden2020}
H.~Weingarden, A.~Matic, R.~G. Calleja, J.~L. Greenberg, O.~Harrison, and S.~Wilhelm.
\newblock Optimizing smartphone-delivered cognitive behavioral therapy for body dysmorphic disorder using passive smartphone data: Initial insights from an open pilot trial.
\newblock {\em JMIR MHealth and UHealth}, 8:Article 6, 2020.

\bibitem{Enander2016}
J.~Enander, E.~Andersson, D.~Mataix-Cols, L.~Lichtenstein, K.~Alstr{\"o}m, G.~Andersson, and C.~R\"uck.
\newblock Therapist guided internet based cognitive behavioural therapy for body dysmorphic disorder: Single blind randomised controlled trial.
\newblock {\em BMJ (Online)}, 4:e005923, 2016.

\bibitem{Wilhelm2022}
S.~Wilhelm, H.~Weingarden, J.~L. Greenberg, S.~S. Hoeppner, I.~Snorrason, E.~E. Bernstein, T.~H. McCoy, and O.~T. Harrison.
\newblock Efficacy of app-based cognitive behavioral therapy for body dysmorphic disorder with coach support: Initial randomized controlled clinical trial.
\newblock {\em Psychotherapy and Psychosomatics}, 91:277--3285, 2022.

\bibitem{Greenberg2022}
J.~L. Greenberg, N.~C. Jacobson, S.~S. Hoeppner, E.~E. Bernstein, I.~Snorrason, A.~Schwartzberg, G.~Steketee, K.~A. Phillips, and S.~Wilhelm.
\newblock Early response to cognitive behavioral therapy for body dysmorphic disorder as a predictor of outcomes.
\newblock {\em Journal of Psychiatric Research}, 152:7313, 2022.
\newblock PMCID: PMC9447469.

\bibitem{Flygare2020}
O.~Flygare, J.~Enander, E.~Andersson, B.~Lj{\'o}tsson, V.~Z. Ivanov, D.~Mataix-Cols, and C.~R\"uck.
\newblock Predictors of remission from body dysmorphic disorder after internet-delivered cognitive behavior therapy: a machine learning approach.
\newblock {\em BMC Psychiatry}, 20:247, 2020.

\bibitem{molnar2019interpretable}
Christoph Molnar.
\newblock {\em Interpretable Machine Learning}.
\newblock Leanpub, 2019.

\bibitem{Devilly2000}
G.~J. Devilly and T.~D. Borkovec.
\newblock Psychometric properties of the credibility/expectancy questionnaire.
\newblock {\em Journal of Behavior Therapy and Experimental Psychiatry}, 31(2):73--86, 2000.

\bibitem{Arch2015}
J.~J. Arch, M.~P. Twohig, B.~J. Deacon, L.~N. Landy, and E.~J. Bluett.
\newblock The credibility of exposure therapy: Does the theoretical rationale matter?
\newblock {\em Behaviour Research and Therapy}, 72:81--392, 2015.

\bibitem{Constantino2018}
M.~J. Constantino, A.~E. Coyne, J.~F. Boswell, B.~R. Iles, and A.~V{\^i}sl{\^a}.
\newblock A meta-analysis of the association between patients' early perception of treatment credibility and their posttreatment outcomes.
\newblock {\em Psychotherapy (Chicago, Ill.)}, 55(4):486--495, 2018.

\bibitem{Bernstein2023}
E.~E. Bernstein, H.~Weingarden, J.~L. Greenberg, J.~Williams, S.~S. Hoeppner, I.~Snorrason, K.~A. Phillips, O.~Harrison, and S.~Wilhelm.
\newblock Credibility and expectancy of smartphone-based cognitive behavioral therapy among adults with body dysmorphic disorder.
\newblock {\em Journal of Obsessive-Compulsive and Related Disorders}, 36, 2023.

\bibitem{Rubin1976}
D.~B. Rubin.
\newblock Inference and missing data.
\newblock {\em Biometrika}, 63:581--592, 1976.

\bibitem{Rubin2004}
D.~B. Rubin.
\newblock {\em Multiple Imputation for Nonresponse in Surveys}.
\newblock John Wiley \& Sons, Inc., 2004.

\bibitem{James2013}
G.~James, D.~Witten, T.~Hastie, and R.~Tibshirani.
\newblock {\em An Introduction to Statistical Learning}, volume 112.
\newblock Springer, 2013.

\bibitem{Sheehan1998}
D.~V. Sheehan, Y.~Lecrubier, K.~H. Sheehan, P.~Amorim, J.~Janavs, E.~Weiller, T.~Hergueta, R.~Baker, and G.~C. Dunbar.
\newblock The mini-international neuropsychiatric interview (m.i.n.i.): The development and validation of a structured diagnostic psychiatric interview for dsm-iv and icd-10.
\newblock {\em The Journal of Clinical Psychiatry}, 59(Suppl 20):22--33, quiz 34--57, 1998.

\bibitem{Harris2009}
P.~A. Harris, R.~Taylor, R.~Thielke, J.~Payne, N.~Gonzalez, and J.~G. Conde.
\newblock Research electronic data capture (redcap): A metadata-driven methodology and workflow process for providing translational research informatics support.
\newblock {\em Journal of Biomedical Informatics}, 42:Article 2, 2009.

\bibitem{Greenberg2019}
J.~L. Greenberg, K.~A. Phillips, G.~Steketee, S.~S. Hoeppner, and S.~Wilhelm.
\newblock Predictors of response to cognitive-behavioral therapy for body dysmorphic disorder.
\newblock {\em Behavior Therapy}, 50:839--849, 2019.
\newblock PMCID: PMC6582981.

\bibitem{Fernandez2021}
L.~Fern\'andez de~la Cruz, J.~Enander, C.~R\"uck, S.~Wilhelm, K.~A. Phillips, G.~Steketee, S.~Sarvode~Mothi, G.~Krebs, L.~Bowyer, B.~Monzani, D.~Veale, and D.~Mataix-Cols.
\newblock Empirically defining treatment response and remission in body dysmorphic disorder.
\newblock {\em Psychological Medicine}, 51:833--839, 2021.

\bibitem{Bowen2016}
D.~Bowen, J.~Jabson, and C.~Kamen.
\newblock mhealth: an avenue for promoting health among sexual and gender minority populations?
\newblock {\em mHealth}, 2:36, 2016.

\bibitem{Gilbey2020}
D.~Gilbey, H.~Morgan, A.~Lin, and Y.~Perry.
\newblock Effectiveness, acceptability, and feasibility of digital health interventions for lgbtiq+ young people: Systematic review.
\newblock {\em Journal of Medical Internet Research}, 22:e20158, 2020.

\bibitem{Lucassen2013}
M.~F.~G. Lucassen, S.~Hatcher, K.~Stasiak, T.~Fleming, M.~Shepherd, and S.~N. Merry.
\newblock The views of lesbian, gay and bisexual youth regarding computerised self-help for depression: An exploratory study.
\newblock {\em Advances in Mental Health}, 12:223--233, 2013.

\bibitem{McGuire2008}
T.~G. McGuire and J.~Miranda.
\newblock New evidence regarding racial and ethnic disparities in mental health: Policy implications.
\newblock {\em Health Affairs}, 27:393--403, 2008.

\bibitem{Phillips2014}
K.~A. Phillips, A.~Hart, and W.~Menard.
\newblock Psychometric evaluation of the yale-brown obsessive-compulsive scale modified for body dysmorphic disorder (bdd-ybocs).
\newblock {\em Journal of Obsessive-Compulsive and Related Disorders}, 3:205--208, 2014.

\bibitem{McConnaughy1983}
E.~A. McConnaughy, J.~O. Prochaska, and W.~F. Velicer.
\newblock Stages of change in psychotherapy: Measurement and sample profiles.
\newblock {\em Psychotherapy: Theory, Research, and Practice}, 20:368--375, 1983.

\bibitem{Eisen1998}
J.~Eisen, K.~A. Phillips, L.~Baer, D.~Beer, K.~Atala, and S.~Rasmussen.
\newblock Brown assessment of beliefs scale: Reliability and validity.
\newblock {\em American Journal of Psychiatry}, 155:102--108, 1998.

\bibitem{Phillips2013}
K.~A. Phillips, A.~S. Hart, W.~Menard, and J.~L. Eisen.
\newblock Psychometric evaluation of the brown assessment of beliefs scale in body dysmorphic disorder.
\newblock {\em The Journal of Nervous and Mental Disease}, 201:640--643, 2013.

\bibitem{Rush2003}
A.~J. Rush, M.~H. Trivedi, H.~M. Ibrahim, T.~J. Carmody, B.~Arnow, D.~N. Klein, ..., and M.~B. Keller.
\newblock The 16-item quick inventory of depressive symptomatology (qids), clinician rating (qids-c), and self-report (qids-sr): A psychometric evaluation in patients with chronic major depression.
\newblock {\em Biological Psychiatry}, 54:573--583, 2003.

\bibitem{Pedregosa2011}
F.~Pedregosa, G.~Varoquaux, A.~Gramfort, V.~Michel, B.~Thirion, O.~Grisel, M.~Blondel, P.~Prettenhofer, R.~Weiss, V.~Dubourg, J.~Vanderplas, A.~Passos, D.~Cournapeau, M.~Brucher, M.~Perrot, and E.~Duchesnay.
\newblock Scikit-learn: Machine learning in python.
\newblock {\em Journal of Machine Learning Research}, 12:2825--2830, 2011.

\bibitem{Hosmer2013}
D.~W. Hosmer, S.~Lemeshow, and R.~X. Sturdivant.
\newblock {\em Applied Logistic Regression}.
\newblock John Wiley \& Sons, 2013.

\end{thebibliography}

\end{document}